\documentclass[conference]{IEEEtran}
\IEEEoverridecommandlockouts
\usepackage{cite}
\usepackage{amsmath,amssymb,amsfonts}
\usepackage{algorithmic}
\usepackage{graphicx}
\usepackage{textcomp}
\usepackage{xcolor}
\usepackage{amsmath}
\usepackage{amsmath}
\let\oldcdot\cdot
\usepackage{breqn}
\usepackage{array}
\usepackage{multirow}
\usepackage{braket}
\usepackage{booktabs}
\usepackage{url} 

\let\cdot\oldcdot
\def\BibTeX{{\rm B\kern-.05em{\sc i\kern-.025em b}\kern-.08em
    T\kern-.1667em\lower.7ex\hbox{E}\kern-.125emX}}
    
\author{\IEEEauthorblockN{Khaled A.~Helal Kelany\IEEEauthorrefmark{1}\IEEEauthorrefmark{2},
Nikitas Dimopoulos\IEEEauthorrefmark{1},
Clemens P. J.~Adolphs\IEEEauthorrefmark{2},
Bardia Barabadi\IEEEauthorrefmark{1}, and
Amirali Baniasadi\IEEEauthorrefmark{1}}
\IEEEauthorblockA{\IEEEauthorrefmark{1}Department of Electrical and Computer Engineering\\
University of Victoria, Victoria, B.C., Canada}
\IEEEauthorblockA{\IEEEauthorrefmark{2}1QB Information Technologies (1QBit)\\
Vancouver, B.C., Canada}
\IEEEauthorblockA{{khaled.kelany@1qbit.com},
{nikitas@ece.uvic.ca},
{clemens.adolphs@1qbit.com},
{bardiabarabadi@uvic.ca},
{amirali@ece.uvic.ca}}}

\begin{document}

\title{Quantum Annealing Approaches to the Phase-Unwrapping Problem in Synthetic-Aperture Radar Imaging  
\thanks{This research was supported in part by a Mitacs Accelerate internship.}
}

\maketitle

\begin{abstract}
The focus of this work is to explore the use of quantum annealing solvers for the problem of phase unwrapping of synthetic aperture radar (SAR) images.
Although solutions to this problem exist based on network programming, these techniques do not scale well to larger-sized images.
Our approach involves formulating the problem as a quadratic unconstrained binary optimization (QUBO) problem, which can be solved using a quantum annealer.
Given that present embodiments of quantum annealers remain limited in the number of qubits they possess, we decompose the problem into a set of subproblems that can be solved individually. These individual solutions are close to optimal up to an integer constant, with one constant per sub-image.
In a second phase, these integer constants are determined as a solution to yet another QUBO problem.
We test our approach with a variety of software-based QUBO solvers and on a variety of images, both synthetic and real. Additionally, we experiment using D-Wave Systems’s quantum annealer, the D-Wave 2000Q. 
The software-based solvers obtain high-quality solutions comparable to state-of-the-art phase-unwrapping solvers. We are currently working on optimally mapping the problem onto the restricted topology of the quantum annealer to improve the quality of the solution.
\end{abstract}

\begin{IEEEkeywords}
Quantum Annealing, Phase Unwrapping, QUBO.
\end{IEEEkeywords}

\section{Introduction}
Two-dimensional phase unwrapping is the process of recovering unambiguous phase values from a two-dimensional array of phase values known only modulo $2\pi$ rad.
The measured phase is also affected by random noise and systematic distortions. 
This problem arises when the phase is used as a proxy indicator of a physical quantity, which is the time delay between two signals in the case of interferometric synthetic aperture radar (InSAR)~\cite{curlander1991synthetic}. 
This time delay is significant, as it is affected by the height differences of the illuminated target. 
It can thus be used to extract accurate three-dimensional topography and reveal topographical changes that occur over time.
As the phase is observable only on a circular space where all measured values are mapped to the range $(-\pi, \pi]$, 
the observed data must be mapped back to the full range of real phase values in order to be meaningful.
For unwrapping purposes, the sampling rate is typically assumed to be suitable for most datasets to prevent aliasing. That is, the absolute difference in phase between two adjacent data points is assumed to be smaller than $\pi$.
This phase-unwrapping problem represents a class of imaging techniques that include InSAR, magnetic resonance imaging, and optical interferometry.

The development of InSAR and many other applications has stimulated interest in building accurate two-dimensional phase-unwrapping algorithms.
The most commonly used unwrapping technique is based on network programming strategies that formulate the problem as a minimum cost flow (MCF)~\cite{costantini1998novel} problem. 
One of these solvers is the sequential tree-reweighted message passing (TRWS) algorithm~\cite{kolmogorov2006convergent}. 
However, since the InSAR images can be quite large---normally larger than 60 M pixels---the process of phase unwrapping via TRWS can take a prohibitively long time on a classical computer.
Thus, we explore whether a quantum computing system could be a potential candidate for solving the phase-unwrapping problem.

Quantum computing exploits the laws of quantum mechanics to process information~\cite{nielsen2002quantum}. 
In contrast to classical computers, which use bits to process information, quantum computers use quantum bits, or qubits, as the basic units of quantum information. 
Analogously to bits, qubits encode state information. 
Qubits may be in either of the two distinct states of $\Ket0$ or $\Ket1$, but they may also encode a superposition of these states, (i.e., $\alpha\Ket0+\beta\Ket1$, with complex-valued coefficients 
$\alpha$ and $\beta$).

Quantum annealing is a quantum computing method used to find the optimal solution to certain combinatorial optimization problems~\cite{finnila1994quantum}.
This is achieved by using properties of quantum mechanics such as quantum tunnelling, entanglement, and superposition.

Quantum annealing systems are able to solve problems in quadratic unconstrained binary optimization (QUBO) form. 
Any unconstrained quadratic integer problem with bounded integer variables can be transformed by a binary expansion into QUBO form~\cite{glover2018tutorial}.
The phase-unwrapping problem is a quadratic unconstrained problem by default, and it can be mapped to a QUBO problem by simply encoding each variable (e.g., $k_{t}$ in~\eqref{eq:6}) into a vector of binary variables.

Due to the limitations in accessing actual quantum annealing infrastructure, we have tested our methodology using a variety of QUBO solvers.
As we will show, the results we obtain match the results obtained using the classical network optimization method (i.e., the TRWS method), which is considered the benchmark in addressing the unwrapping problem.

InSAR images tend to be quite large, often exceeding \mbox{20 k $\times$ 30 k}, or \mbox{600 M}, pixels. In the simplest problem where each pixel label would require one qubit, a 600-M-qubit quantum annealer would be required; such a machine is not currently available. 
To overcome the limitations of present-day technology, we have developed a method where we partition an image and then use quantum annealing on the individual partitions to obtain suboptimal labelling, after which we use quantum annealing in a second phase to obtain labels that approach the ones obtained through classical methods.
We have named our method “super-pixel decomposition". 

The rest of the paper is organized as follows. 
In Section~\ref{section:Background} we provide background information, 
in Section~\ref{section:Methodology} we explain our methodology, 
and in Section~\ref{section:Experiments} we present our experimental results. We give our conclusion in Section~\ref{section:Conclusion}.

\section{Background}\label{section:Background}

\subsection{Phase-unwrapping formulation}

Strictly speaking, phase unwrapping is an ill-posed problem, as the
unwrapped phase array contains information that is not available in
the wrapped array. Therefore, to perform correctly, all phase-unwrapping methods rely on
regularizing assumptions. The most common of these assumptions
is that the Nyquist criterion is met throughout most (but not necessarily
all) of the scene. That is, the spatial sampling rate is assumed to be
high enough that aliasing is avoided~\cite{chen2002phase}.

The Nyquist criterion implies that the difference between the phases of
two neighbouring pixels is less than $2\pi$. The key to phase unwrapping,
therefore, lies not on directly calculating the unwrapped phase values
themselves, but in estimating these values given that the differences
of the wrapped phases is the same as those of the unwrapped phases
dictated by the Nyquist assumption.

Let $\phi$, $\varphi$, and $k$ denote the unwrapped phase, the wrapped
phase, and an integer label to be estimated, respectively. For the phase of a pixel $i$, we have 

\begin{equation}
\label{eq:1}
\phi_{i}=\varphi_{i}+2\pi k_{i}\,.
\end{equation}

The unwrapping problem can then be expressed as an optimization problem
of the cost function
\begin{equation}
E=\sum_{(s,t)\text{\ensuremath{\in}}A}W_{st}|k_{t}-k_{s}-a_{st}|\,,
\label{eq:2}
\end{equation}
that is, 
\begin{equation}
\text{argmin}_{k}=\sum_{(s,t)\text{\ensuremath{\in}}A}W_{st}|k_{t}-k_{s}-a_{st}|\,,
\end{equation}
where $k_{i}$ are the labels that will determine the original phase
as per~\eqref{eq:1}, $A$ is the set of pixels in the SAR image, $W_{st}$
are weights defining the neighbourhood structure, and $a_{ij}$ are
constants obtained from the image as per the equation

\begin{equation}
a_{ij}\stackrel{\text{def}}{=}\frac{\text{wrap}\left(\phi_{i}-\phi_{j}\right)-\left(\phi_{i}-\phi_{j}\right)}{2\pi},
\label{eq:3}
\end{equation}
where 
\begin{equation}
\text{wrap}\left(\theta\right)=\text{arg}\left(e^{i\theta}\right)=\theta-\lfloor\frac{\theta}{2\pi}\rfloor\,.
\end{equation}
The Appendix details the derivation of the cost function shown in \eqref{eq:2}.

The optimization problem as defined in \eqref{eq:3} above admits several solutions,
as only the difference of the labels is used in the cost function.
Labels can be increased or decreased by the same amount and still
result in the same minimal cost. A way to further regularize the solution is to insist that the desirable solution involve labels that are the smallest ones possible. Therefore, the cost function is augmented with an extra term that depends on the labels themselves as follows: 

\begin{equation}
E=\sum_{(s,t)\text{\ensuremath{\in}}A}W_{st}|k_{t}-k_{s}-a_{st}|+\sum_{s\text{\ensuremath{\in}}A}\omega_{s}|k_{s}-a_{s}|
\label{eq:6}
\end{equation}

The weights $W_{st}$, $\omega_{s}$, and the bias $a_{s}$, are chosen
heuristically and represent ad hoc information one may have on the
scene represented in the image, most often $a_{s}=0$. 

Without loss of generality, one can also consider cost functions involving
quadratic expressions of the labels instead of the more challenging
absolute value ones,

\begin{equation}
E=\sum_{(s,t)\text{\ensuremath{\in}}A}W_{st}\left(k_{t}-k_{s}-a_{st}\right)^{2}+\sum_{s\text{\ensuremath{\in}}A}\omega_{s}\left(k_{s}-a_{s}\right)^{2},
\label{eq:phase_uwrapping_old}
\end{equation}
and in the case that $a_{s}=0$,  a similar cost function is 
\begin{equation}
E=\sum_{(s,t)\text{\ensuremath{\in}}A}W_{st}\left(k_{t}-k_{s}-a_{st}\right)^{2}+\sum_{s\text{\ensuremath{\in}}A}\omega_{s}k_{s}^{2}\,.
\label{eq:phase_uwrapping}
\end{equation}

Although the $L_1$-norm is preferred over the $L_2$-norm in the continuous case---as the $L_2$-norm tends to spread the error and does not result in good
solutions~\cite{costantini1998novel}---this is not a factor in the integer case.
The most commonly used method
of solving the phase-unwrapping problem (TRWS) is attributed to \mbox{V.
Kolmogorov~\cite{kolmogorov2006convergent}.} 

\subsection{Quantum Annealing}
Quantum annealing employs quantum tunnelling to ensure that a system is able to escape local minima as it traverses the state space of an energy function toward its way to ground-state settlement.
Quantum computational systems, such as the ones manufactured by D-Wave Systems, use quantum annealing to locate the ground state of an artificial Ising system~\cite{johnson2011quantum}.
An Ising Hamiltonian describes the behaviour of such a system as

\begin{equation}
H_{p}=\sum_{i=1}^{N}h_{i}\sigma_{i}^{z}+\sum_{i,j=1}^{N}J_{ij}\sigma_{i}^{z}\sigma_{j}^{z}\,,
\label{eq:Hamiltonian}
\end{equation}
where $h_{i}$ is the energy bias for spin $i$, $J_{ij}$ is the coupling energy between spins $i$ and $j$, $\sigma_{i}^{z}$ is the Pauli spin matrix, and $N$ is the number of qubits. 
Quantum annealing on this system is achieved by the gradual evolution of the Hamiltonian system~\cite{johnson2011quantum}
\begin{equation}
H\left(t\right)=\Gamma\left(t\right)\sum_{i=1}^{N}\Delta_{i}\sigma_{i}^{x}+\varLambda\left(t\right)H_{p}\,.
\end{equation}
As time passes, $\Gamma$ decreases from 1 to 0 while $\varLambda$ increases from 0 to 1.
If the annealing process is performed sufficiently slowly, the system remains in the ground state of $H(t)$ for all times, $t$, settling at the end of the annealing process at the ground state of $H_p$.
The Hamiltonian in \eqref{eq:Hamiltonian} can be rewritten in vector form as $H\left(s\right)=\mathbf{s}^{T}\mathbf{Js+\mathbf{s}}^{T}\mathbf{h}$, in the form of a QUBO problem~\cite{zaribafiyan2017systematic}.

As used in the rest of this paper, the objective function is expressed in QUBO form in scalar notation, and is defined as 
\begin{equation}
C\left(x\right)=\sum_{i}a_{i}x_{i}+\sum_{i<j}b_{i,j}x_{i}x_{j}\,,
\end{equation}
where $x\in\left\{ 0;1\right\} ^{n}$ is a vector of binary variables
and $\left\{ a_{i};b_{i;j}\right\} $ are real coefficients. 

Before an application problem can be solved on a quantum annealer, it must first be mapped into QUBO form.
As a first step in transforming the InSAR problem into a QUBO problem, the $k_i$ label that is non-binary valued must be transformed into one that is binary
valued. Let $k_{i}\in\left\{ 0,D_{i}-1\right\}$, where $D_{i}$ is
the number of allowed values (labels) for $k_{i}$. This can be achieved
by writing $k_{i}$ in binary form. The binary transformation restricts the
number of new-valued binary variables required to represent $k_{i}$.
Let $d_{i}=\lceil \text{log}_{2}D_{i}\rceil$ and $k_{i}=\langle\mathbf{2},\mathbf{x_{i}}\rangle$,
where the vector $\mathbf{x}_{\mathbf{i}}=[x_{i,d_{i}},\cdots,x_{i,1},x_{i,0}]$
represents the bits of $k_{i}$ and $\mathbf{2}=[2^{d_{i}},\cdots,2,1]$
is the vector of powers of two. Equation~\eqref{eq:phase_uwrapping_old} can be written in QUBO
form as
\begin{dmath}
E=\sum_{(s,t)\text{\ensuremath{\in}}A}W_{st}\left(\sum_{i}b_{i}x_{i,t}\text{\textminus}\sum_{i}b_{i}x_{i,s}\text{\textminus}a_{st}\right)^{2}+\sum_{s\text{\ensuremath{\in}}A}\omega_{s}\left(\sum_{i}b_{i}x_{i,s}\text{\textminus}a_{s}\right)^{2},
\end{dmath}
where $b_{i}$ is the weighting coefficient for the binary variable $x_i$ ($b_{i}=2^i$ in the case of the binary encoding). 

Many problems can be formulated to take advantage of quantum annealing, which is advantageous because it converges faster than other techniques to an optimum solution~\cite{jooya2017accelerating}.

Quantum annealing can be compared to simulated annealing by identifying that the temperature parameter in simulated annealing performs a similar role to quantum tunnelling in quantum annealing. 
The temperature in simulated annealing defines the probability of moving from a single current state to a higher energy state to escape local minima. 
The assumed advantage of quantum annealing over simulated annealing is that tunnelling allows the system to directly pass through high energy barriers without having to climb over them.

Analytical and numerical evidence indicates that quantum annealing can outperform simulated annealing~\cite{heim2015quantum}. Therefore,
quantum annealing is a good potential solver for the InSAR phase-unwrapping problem. 

\subsection{Optimizers}

As mentioned earlier, the size of the problem prevents the direct use of currently available annealing infrastructure. 
Similarly, the size of the problem results in a prohibitively expensive QUBO computation if one elects to perform a global optimization on the full-scale image.
Thus, the methodology involves partitioning an image, and then applying QUBO solvers first on the partitions and then in a second phase on an abstraction of the image comprising what we call ``super-pixels'', each one representing a partition of the original image.

In the following sections, (\ref{subsec:classical_optimzer} and \ref{subsec:qubo_optimzer}), we discuss the QUBO solvers we have employed, after which we discuss our image partitioning approach and the two-phase super-pixel methodology.

\subsubsection{Classical Optimizer}\label{subsec:classical_optimzer}

\paragraph{TRWS~\emph{\cite{kolmogorov2006convergent}}} 

The TRWS algorithm is used for discrete energy minimization, where the energy function can be formulated as

\begin{equation}
E\left(x|\theta\right)=\theta_{\text{const}}+\sum_{s\in\nu}\theta_{s}\left(x_{s}\right)+\sum_{(s,t)\in\varepsilon}\theta_{st}\left(x_{s},x_{t}\right),
\end{equation}
where $\nu$ corresponds to the set of pixels, $x_{s}$ indicates the label of pixel $s \in \nu$, 
$\varepsilon$ corresponds to the set of edges (each edge connects two related pixels), 
$\theta_{s}(\cdot)$  is the penalty function (i.e., a term of an unconstrained objective function added to add some constraint to it) of unary data, and
$\theta_{st}(\cdot,\cdot)$ is the penalty function of the pairwise terms.
This energy function is usually derived in the context of Markov random fields~\cite{geman1984stochastic}.
The algorithm is widely used in phase unwrapping problems, where the unary penalty functions represent the unary terms in~\eqref{eq:phase_uwrapping_old}, where the pixels are penalized for having large values, while the pairwise penalty functions represent the pairwise terms in~\eqref{eq:phase_uwrapping_old}, where the two pixels $k_t$ and $k_s$ are penalized for having a difference not equal to $a_{st}$. 

\subsubsection{QUBO Optimizer}\label{subsec:qubo_optimzer}

\paragraph{PTICM} 

Parallel tempering with isoenergetic cluster moves (PTICM) is one of the parallel tempering algorithms introduced in~\cite{zhu2020borealis}, which is a Monte Carlo approach for solving QUBO problems.
The PTICM algorithm simultaneously simulates multiple replicas of the original system at different temperatures. Each of the replicas has a different initial state.
The replicas are regularly swapped with neighbouring temperatures based on a Metropolis criterion.
These swaps enable the different replicas to make a random walk in the temperature space, allowing the efficient overcoming of energy barriers.

\paragraph{Parallel Tempering}
We experimented using an alternative implementation of parallel tempering. This solver gives accuracy similar to, if not better than, PTICM. This solver is one of Microsoft's Quantum Inspired Optimization (Microsoft QIO) solvers, and is accessible through a cloud--client interface. However, at present we have limited early access to this solver. Hence, we made use of it only on small images.

\paragraph{Simulated Annealing}
This solver provides an implementation of the simulated annealing method~\cite{swendsen1986replica}. The solver is also one of the Microsoft QIO solvers. Therefore, we used this solver on small images for the same reason mentioned above.

\paragraph{D-Wave Annealing}
D-Wave Systems provides implementations of different quantum annealing systems, starting from the \mbox{D-Wave One} announced in 2011~\cite{johnson2011quantum}. 
We used the \mbox{D-Wave 2000Q\_6} machine to unwrap the InSAR sub-images. 
The machine contains 2041 qubits. 
The qubits are sparsely connected in an architecture known as a ``Chimera'' graph. The Chimera architecture comprises sets of connected unit cells.
Each unit cell has four horizontal qubits that are connected via couplers to four vertical qubits.
Unit cells are tiled horizontally and vertically with adjacent qubits connected. 
The qubits are logically mapped into a matrix of 16 $\times$ 16 unit cells, with eight qubits per cell. 
In theory, the Chimera architecture comprises $16 \times 16 \times 8=2048$ qubits.
In practice, however, the largest number of embeddable qubits is slightly smaller (2041 qubits) due to missing, or faulty, qubits, an issue that arises during  manufacturing.
This also results in there being some nonexistent connections.

\section{Methodology}\label{section:Methodology}
In this section, we describe our methodology in breaking down the phase-unwrapping problem into smaller problems that are easier to solve. 
The smaller problems can be solved in parallel and approach the global solution.

\subsection{Phase-Unwrapping Decomposition}

Equation~\eqref{eq:phase_uwrapping_old} describes the energy function for
phase unwrapping. The equation consists of two terms: the pair-wise
term $W_{st}\left(k_{t}-k_{s}-a_{st}\right)^{2}$ that describes the relationship
between two pixels $s$ and $t$, and the unary term $\omega_{s}\left(k_{s}-a_{s}\right)^{2}$
that describes the energy of the pixel $s$. 

Phase-unwrapping decomposition is based on a divide-and-conquer
approach. A given large InSAR image is subdivided into smaller sub-images
that fit onto a quantum annealing machine. Each of these sub-images
are unwrapped independently. The sub-images are then stitched together to form the final unwrapped large InSAR image.

This approach introduces a problem at the boundaries of the sub-images.
Since we are unwrapping the sub-images separately, there is no guarantee
that the labelling of two pixels, each belonging to a boundary of two adjacent sub-images, will attain optimal labelling consistent
with that which would be obtained if both pixels had been part of the same optimization
problem (i.e., if we had unwrapped the entire image all at once). 

We assume that the pixels in each sub-image are labelled correctly
up to an integer additive factor, where all the pixels of one sub-image share the same additive factor. 
Then, our objective is to find
those additive factors such that all the pixels at the boundaries will
be consistent.

We propose a super-pixel heuristic to determine these additive factors as we now will describe.

The wrapped InSAR image is divided into non-overlapping sub-images,
where each sub-image contains a subset of the pixels. 
The energy function determined by \eqref{eq:phase_uwrapping_old} can be rewritten as
\begin{dmath}
E=
\sum_{g\in G}\left[\sum_{(s,t)\text{\ensuremath{\in}}A_{g}}W_{st}\left(k_{t}-k_{s}-a_{st}\right)^{2}+\sum_{s\text{\ensuremath{\in}}A_{g}}\omega_{s}\left(k_{s}-a_{s}\right)^{2}\right]\\ +
\sum_{\begin{array}{c}
t\in A_{i},\\
s\in A_{j},\\
i\neq j\,,
\end{array}}W_{st}\left(k_{t}-k_{s}-a_{st}\right)^{2},
\label{eq:8}
\end{dmath}
where $G$ is the set of the sub-images, and $A_{x}$ is the set of
pixels in the sub-image $x$. 

In the above equation, the terms within the square brackets correspond to an energy function for each sub-image while the last sum collects all the terms that connect the sub-images.
Our approach is to optimize each of the sub-images, that is, to determine the labels that optimize the energy functions corresponding to each sub-image separately.
We assume next that the obtained solutions are correct---that they are identical plus or minus a sub-image wide integer shift---to the solution obtained when we optimize the image in its totality.
The next step is to determine these additive factors, which can be formulated as a QUBO problem.
Let $K_s$ denote the additive factor corresponding to sub-image $s$, and let $k_{i}^{'}$ denote the label of pixel $i$ as determined by the QUBO formulation of each sub-image.
Then $k = k_{i}^{'} + K_s$ for $i$ in $A_s$.

Equation~\eqref{eq:8} can now be rewritten as follows:

\begin{dmath}
E=
\sum_{g\text{\ensuremath{\in}}G}\left[
\sum_{(s,t)\text{\ensuremath{\in}}A_{g}}
W_{st}\left(k_{t}^{'}+K_{g}-k_{s}^{'}-K_{g}\right)^{2}+
\sum_{s\text{\ensuremath{\in}}A_{g}}\omega_{s}\left(k_{s}^{'}+K_{g}-a_{s}\right)^{2}\right]+
{\displaystyle \sum_{\begin{array}{c}
t\in A_{i},\\
s\in A_{j},\\
i\neq j
\end{array}}W_{st}\left(k_{t}^{'}+K_{i}-k_{s}^{'}-K_{j}-a_{st}\right)^{2}},
\label{eq:9}
\end{dmath}
The first sum within the bracket is devoid of $K_g$ and it is constant as the labels $k_s^{'}$ have been determined by the previous QUBO operation. Ignoring constant terms, Equation~\eqref{eq:9} can be rewritten as
\begin{dmath}
\tilde{E}=
\sum_{g\text{\ensuremath{\in}}G}\left[
\sum_{s\text{\ensuremath{\in}}A_{g}}\omega_{s}\left(k_{s}^{'}+K_{g}-a_{s}\right)^{2}, 
\right]+
{\displaystyle \sum_{\begin{array}{c}
t\in A_{i},\\
s\in A_{j},\\
i\neq j.
\end{array}}
W_{st}\left(\left(k_{t}^{'}-k_{s}^{'}\right)+
\left(K_{i}-K_{j}\right)-a_{st}\right)^{2}}
\label{eq:10}
\end{dmath}
Denoting $a_s^{'}=a_s-k_s^{'}$ and $a_{st}^{'}=a_{st}-\left(k_t^{'}-k_s^{'}\right)$, Equation~\eqref{eq:10} is written as

\begin{dmath}
\tilde{E}=
\sum_{g\text{\ensuremath{\in}}G}\left[\sum_{s\text{\ensuremath{\in}}A_{g}}\omega_{s}\left(K_{g}-a_{s}^{'}\right)^{2}\right]
+
\sum_{\begin{array}{c}
t\in A_{i},\\
s\in A_{j},\\
i\neq j.
\end{array}}W_{st}\left(K_{i}-K_{j}-a_{st}^{'}\right)^{2},
\label{eq:11}
\end{dmath}
The first term of \eqref{eq:11} is a second-order function, while the second term regularizes the solution by selecting $K_g$ to be as small as possible.
The coefficient $\omega_s$ and the term $a_s$ as per  \eqref{eq:phase_uwrapping_old} and \eqref{eq:phase_uwrapping} have been chosen arbitrarily.
To ensure that our energy function conforms to the form of \eqref{eq:phase_uwrapping_old}  without affecting the accuracy of the solution, we select $\omega_s=\omega_g$ and \mbox{$a_s^{'}=A_g         \ \forall s \in A_g$}. This results in the following expression for the energy function:

\begin{equation}
\hat{E}=\sum_{\begin{array}{c}
t\in A_{i},\\
s\in A_{j},\\
i\neq j
\end{array}}W_{st}\left(K_{i}-K_{j}-a_{st}^{'}\right)^{2}+\sum_{g\text{\ensuremath{\in}}G}\omega_{g}\left(K_{g}-A_{g}\right)^{2},
\end{equation}
This is the energy function of the super-pixel level where $K_g$ represents the sought labels for each sub-image (i.e., super-pixel).
As it is a quadratic unconstrained integer optimization problem, it is amenable to a QUBO-based solution.

\section{Experiments}\label{section:Experiments}

In this section, we present the experimental results for the super-pixel decomposition approach using a classical solver and other QUBO solvers.
Our objective is to attempt to support the assertion that our approach of partitioning the image and formulating the phase-unwrapping problem as a QUBO problem results in good-quality solutions. 
The objectives of the present work have not been to fully quantify the quality of the solutions obtained nor the efficacy of the various solvers involved.

\subsection{Setup}

\subsubsection{Datasets}

There are eight sets of data used in the experiments. They are summarized in Table~\ref{tbl:A summary of the datasets used}. 
The datasets include simulated data that have different levels of noise and also real data.

\begin{table*}
\caption{Summary of the datasets used}
\begin{centering}
\begin{tabular}[tb]
{c>{\centering}p{1.2cm}>{\centering}p{1.2cm}>{\centering}p{1.2cm}>{\centering}p{1.2cm}>{\centering}p{1.2cm}>{\centering}p{1.2cm}>{\centering}p{1.2cm}>{\centering}p{1.2cm}}
\toprule 
Dataset             & \#1                   & \#2                       & \#3                           & \#4           & \#5                   & \#6                   &\#7 &\#8\tabularnewline
\toprule 
Description         & Simulated noise-free  & Simulated with low noise  & Simulated with high noise     & Real data     & Simulated noise-free  & Real data  & Simulated noise-free  & Real data  \tabularnewline
\hline 
\# images           & 10                    & 10                        & 10                            & 1             & 1                     & 1                     & 1                     & 1                     \tabularnewline
\hline 
Image size          & 400$\times$400               & 400$\times$400                   & 400$\times$400                       & 400$\times$400       & 100$\times$100               & 100$\times$100            & 100$\times$100               & 100$\times$100               \tabularnewline
\hline 
Sub-image size      & 20$\times$20               & 20$\times$20                   & 20$\times$20                       & 20$\times$20       & 16$\times$16                 & 16$\times$16           & 10$\times$10         & 10$\times$10          \tabularnewline
\hline 
Signal generator    & Perlin noise          & Perlin noise              & Perlin noise                  & -             & Perlin noise          & -            & Perlin noise          & -          \tabularnewline
\hline 
Noise in SNR        & -                     & 15 db                     & 13 db                         & 8.1 db        & -                     & 8.1 db                  & -                     & 8.1 db                     \tabularnewline
\hline 
Max ambiguity       & 4                     & 4                         & 4                             & 4             & 4                     & 4        & 4                     & 4                     \tabularnewline
\hline 
Connectivity        & Four neighbours  & Four neighbours      & Four neighbours          & Four neighbours  & Four neighbours  & Four neighbours & Four neighbours  & Four neighbours \tabularnewline
\bottomrule
\end{tabular}
\par\end{centering}
\label{tbl:A summary of the datasets used}
\end{table*}

\subsubsection{Solvers}

The four solvers discussed in the background section, namely, TRWS, PTICM, parallel tempering, and simulated annealing solvers, are used to unwrap the images, the sub-images, and the super-pixel image. 
The classical TRWS solver is the one most commonly used in phase unwrapping and; thus, we use it as a reference to evaluate the accuracy of the other solvers. 
The other solvers are QUBO solvers and demonstrate that our proposed methodology works on various solvers.

\subsubsection{Experiments}

We conducted three experiments, naming them simply ``Experiment 1'', ``Experiment 2'', and ``Experiment 3''.
In Experiment 1, the first four datasets are tested and the statistics of each set are reported.
Only two solvers are considered in this experiment, the TRWS solver and the PTICM solver. 
This is due to our having limited access to the other solvers (i.e., Parallel Tempering, Simulated Annealing, and the D-Wave Annealer). 
In Experiment 2, four solvers are used: TRWS, PTICM, Parallel Tempering, and  Simulated Annealing.  
However, only one image per set is considered.
In Experiment 3, the last four datasets are used to test the TRWS solver and the D-Wave Annealer.

\subsubsection{Accuracy metric}
We use the matching percentage of an image with the ground truth as an accuracy metric. 
That is the number of pixels in the images that match the number of pixels in the ground truth divided by the total number of pixels.

\subsection{Results}

This subsection presents a summary of the results of Experiments 1, 2, and 3. 
The image phase-unwrapping process consists of two steps: first, unwrapping sub-images and then constructing and unwrapping the super-pixel images that give us the final unwrapped image.
The average and standard deviation are reported for the matching percentage for the sub-images and images for each dataset, for each solver.

\subsubsection{Experiment 1 Results}

\begin{figure}
\begin{centering}
\includegraphics[width=7cm]{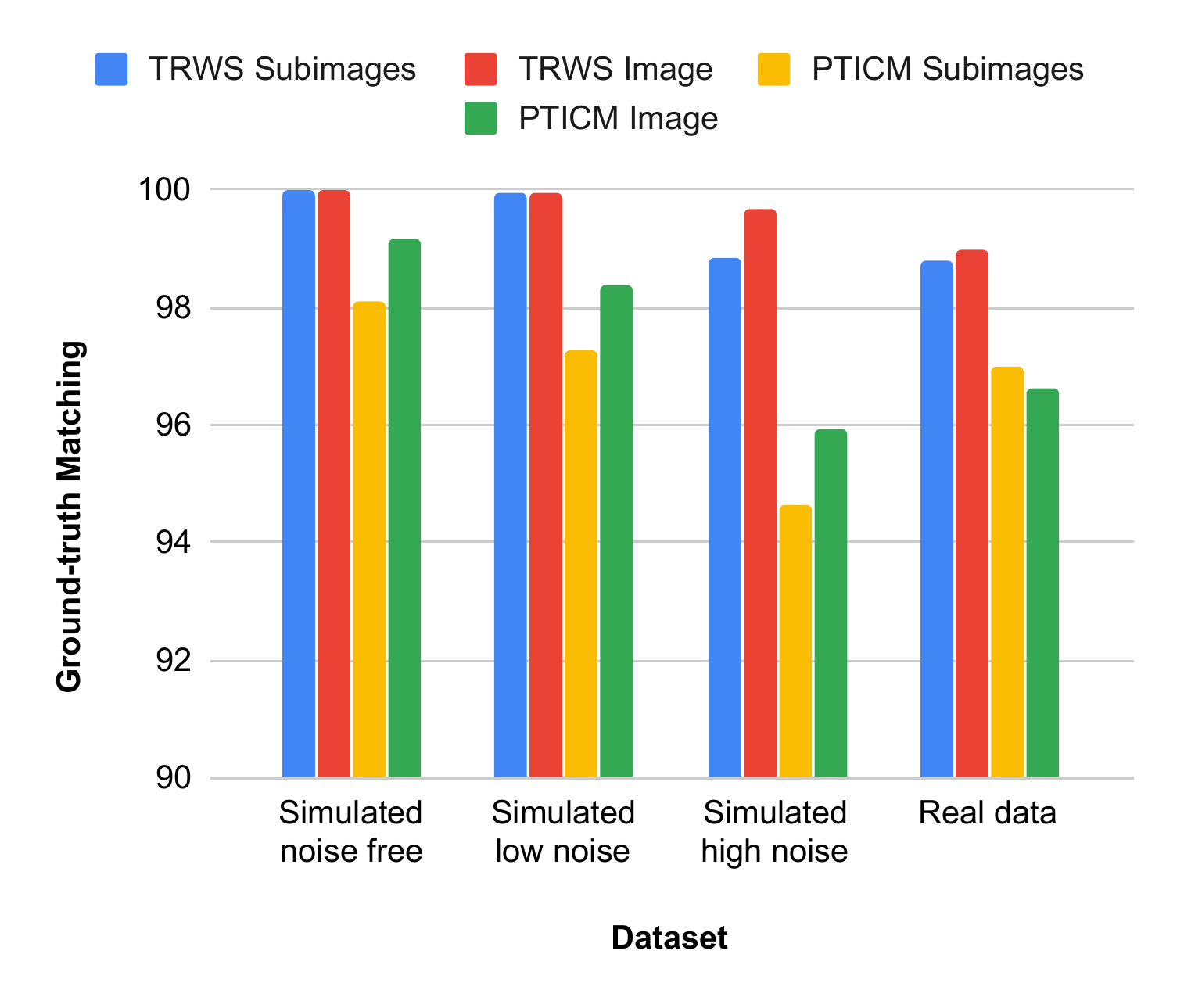}
\par\end{centering}
\caption{Set 1 ground truth matching for the unwrapping of sub-images and the overall
image unwrapping}
\label{fig:se1_results}
\end{figure}

This set of results is for the TRWS and PTICM solvers. Thirty-one images 400 $\times$ 400 in size are used. Thirty images are simulated: 10 are noise-free, 10 have a low level of noise, and 10 have a high level of noise. One image is real. Each image is subdivided into 400 images 20 $\times$ 20 pixels in size. 
The results are summarized in Table~\ref{tbl:set1 results}. 
The last two rows in Table~\ref{tbl:set1 results} present the results for the single real image unwrapped, so there is only one value for the image matching. 
Fig.~\ref{fig:se1_results} shows how close the results produced by the two solvers are to the ground truth.

\begin{table*}
\caption{Results of the unwrapping of sub-images  and the overall image unwrapping
for the TRWS and PTICM solvers}
\begin{centering}
\begin{tabular}{cccccc}
\toprule 
\multirow{2}{*}{Dataset} & \multirow{2}{*}{Solver}  & \multicolumn{2}{c}{Sub-images unwrapping} & \multicolumn{2}{c}{Super-pixel unwrapping}\tabularnewline
                                                    &           & Avg matching (\%)  & STD matching  & Avg matching (\%) & STD matching\tabularnewline
\toprule 
\multirow{2}{*}{\#1 (Noise-Free Simulated Data)}    & TRWS      & 100           & 0             & 100           & 0\tabularnewline
                                                    & PTICM     & 98.1          & 12.000579     & 99.17         & 0.319373\tabularnewline
\hline 
\multirow{2}{*}{\#2 (Low-Noise Simulated Data)}     & TRWS      & 99.96         & 0.206808      & 99.96         & 0.011311\tabularnewline
                                                    & PTICM     & 97.27         & 13.753412     & 98.39         & 1.102349\tabularnewline
\hline 
\multirow{2}{*}{\#3 (High-Noise Simulated Data)}    & TRWS      & 98.82         & 8.503787      & 99.68         & 0.111031\tabularnewline
                                                    & PTICM     & 94.63         & 15.026709     & 95.95         & 1.929165\tabularnewline
\hline 
\multirow{2}{*}{\#4 (Real Data)}                    & TRWS      & 98.79         & 4.968538      & 98.97         & -\tabularnewline
                                                    & PTICM     &  97           & 7.028149      & 96.63         & -\tabularnewline
\bottomrule 
\end{tabular}
\par\end{centering}
\label{tbl:set1 results}
\end{table*}

\subsubsection{Experiment 2 Results}

\begin{figure}
\begin{centering}
\includegraphics[width=7cm]{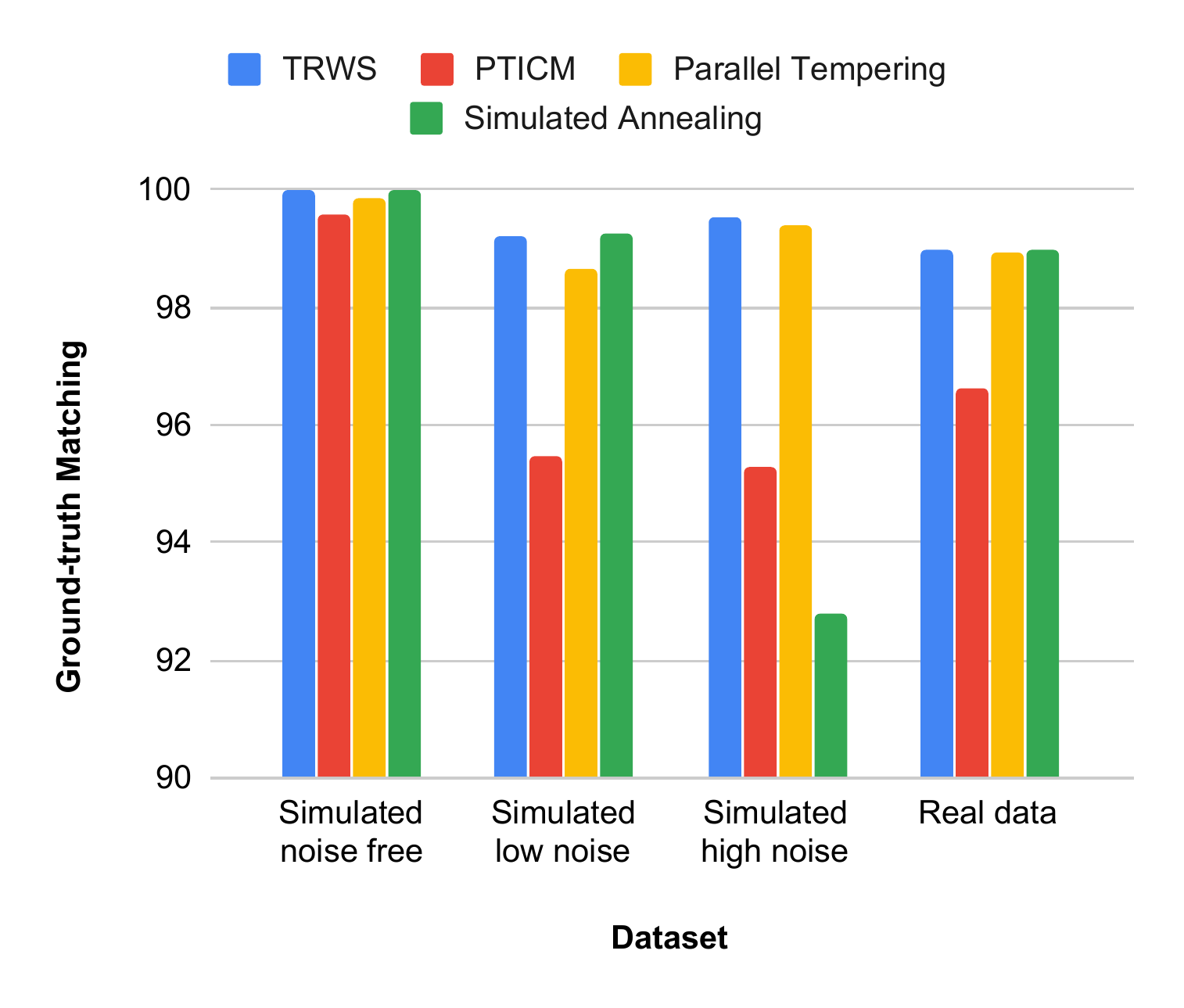}
\par\end{centering}
\caption{Set 2 ground truth matching for the overall image unwrapping}
\label{fig:se2_results}
\end{figure}

This set of results is for the TRWS, PTICM, Parallel Tempering, and Simulated Annealing solvers. 
Only four images 400 $\times$ 400 pixels in size are used in this experiment: one image from each of the datasets \#1, \#2, \#3, and \#4. 
As there is only one image per set, there is only a single value for the ground truth matching, (i.e., there is no average and no standard deviation).
The results are summarized in Table~\ref{tbl:set2_results}. 
Fig.~\ref{fig:se2_results} shows how close the results produced by the solvers are to the ground truth.

\begin{table*}
  \caption{Results of the unwrapping of sub-images and the overall image unwrapping
    for the TRWS, PTICM, Parallel Tempering, and Simulated Annealing solvers}
  \begin{centering}
    \begin{tabular}{ccccc}
      \toprule
      \multirow{2}{*}{Dataset} & \multirow{2}{*}{Solver} & \multicolumn{2}{c}{Sub-images unwrapping} & Super-pixel unwrapping\tabularnewline
                                                       &                     & Avg matching (\%) & STD matching & Matching (\%)\tabularnewline
      \toprule
      \multirow{4}{*}{\#1 (Noise-Free Simulated Data)} & TRWS                & 100        & 0            & 100\tabularnewline
                                                       & PTICM               & 99.6         & 0.638959     & 99.60\tabularnewline
                                                       & Parallel Tempering  & 99.37        & 7.033106     & 99.87\tabularnewline
                                                       & Simulated Annealing & 100        & 0            & 100\tabularnewline
      \hline
      \multirow{4}{*}{\#2 (Low-Noise Simulated Data)}  & TRWS                & 99.24        & 1.764139     & 99.21\tabularnewline
                                                       & PTICM               & 96.64        & 6.240565     & 95.45\tabularnewline
                                                       & Parallel Tempering  & 98.25        & 9.778803     & 98.67\tabularnewline
                                                       & Simulated Annealing & 98.55        & 8.438345     & 99.24\tabularnewline
      \hline
      \multirow{4}{*}{\#3 (High-Noise Simulated Data)} & TRWS                & 100          & 0            & 99.54\tabularnewline
                                                       & PTICM               & 98.64        & 9.862505     & 95.29\tabularnewline
                                                       & Parallel Tempering  & 98.15        & 11.2803      & 99.41\tabularnewline
                                                       & Simulated Annealing & 93.97        & 17.841954    & 92.78\tabularnewline
      \hline
      \multirow{4}{*}{\#4 (Real Data)}                 & TRWS                & 98.79        & 4.968538     & 98.97\tabularnewline
                                                       & PTICM               & 97           & 7.028149     & 96.63\tabularnewline
                                                       & Parallel Tempering  & 98.75        & 1.084935     & 98.95\tabularnewline
                                                       & Simulated Annealing & 99.24        & 1.310232     & 98.96\tabularnewline
      \bottomrule
    \end{tabular}
    \par\end{centering}
  \label{tbl:set2_results}
\end{table*}

\subsubsection{Experiment 3 Results}
This experiment set is designed to test our approach on the D-Wave quantum annealing system, the D-Wave 2000Q. This set of results involve the TRWS and D-Wave 2000Q solvers. Two images are used in this experiment: one simulated noise-free image with a maximum ambiguity of 4, and one real image with a maximum ambiguity of 4. Each image is subdivided into sub-images 16 $\times$ 16 pixels and 10 $\times$ 10 pixels in size to form in four datasets in total. The four sets are labelled dataset \#5, \#6, \#7, and \#8 in Table~\ref{tbl:A summary of the datasets used}. 

Solving a QUBO problem on a quantum annealer requires embedding (mapping) each binary variable to one physical qubit or multiple chained (connected) qubits. The embedding process varies based on the problem and the architecture of a given quantum annealer. 
The D-Wave 2000Q annealer comprises 2048 qubits organized in a 16 $\times$ 16 grid of eight-qubit unit cells.
Each qubit is connected to six neighbours in the Chimera topology~\cite{DWaveQPU21:online}.
However, faulty qubits curtail the connectivity and the size of the problems the machine can handle.
D-Wave Systems provides a tool that heuristically embeds the binary variables to the quantum annealer's qubits. The tool does not provide the optimal embedding in terms of keeping more related qubits closer to each other. 
Hence, we used a manual approach to embed the logical binary variables onto physical qubits.

In manually embedding our variables onto qubits, we strived to produce a symmetric embedding. The symmetry of the embedding contributed to an improved solution quality.
Further, we mapped each integer label to a single Chimera cell.
The largest image our embedding approach can map onto the D-Wave 2000Q is 16 $\times$ 16 pixels in size.
Bypassing the faulty qubits resulted in asymmetries, which lowered the quality of the solutions.
To avoid such issues, we have experimented with smaller images 10 $\times$ 10 pixels in size which, when mapped, avoid the faulty qubits with a concomitant increase in the quality of the solution.
The results are summarized in Table~\ref{tbl:set3 results}. 

\begin{table*}
  \caption{Results of the unwrapping of the sub-images  and the overall image unwrapping
    for the TRWS and D-Wave solvers}
  \begin{centering}
    \begin{tabular}{cccccc}
      \toprule
      \multirow{2}{*}{Dataset} & \multirow{2}{*}{Sub-image size} & \multirow{2}{*}{Solver} & \multicolumn{2}{c}{Sub-images unwrapping} & Avg matching\tabularnewline
                               &                                 &                         & Avg matching  (\%)& STD matching & normalized to TRWS (\%)\tabularnewline
      \toprule
      \multirow{2}{*}{\#5 (Noise-Free Simulated Data)} & \multirow{2}{*}{16$\times$16} & TRWS & 100          & 0            & 100\tabularnewline
                                                       &                        & D-Wave & 85.47        & 16.52        & 85.47\tabularnewline
      \hline
      \multirow{2}{*}{\#6 (Real Data)}                 & \multirow{2}{*}{16$\times$16} & TRWS   & 87.5         & 2.36         & 100\tabularnewline
                                                       &                        & D-Wave & 87.42        & 2.46        & 99.91\tabularnewline
      \hline
      \multirow{2}{*}{\#7 (Noise-Free Simulated Data)} & \multirow{2}{*}{10$\times$10} & TRWS & 100          & 0            & 100\tabularnewline
                                                       &                        & D-Wave & 100        & 0        & 100\tabularnewline
      \hline
      \multirow{2}{*}{\#8 (Real Data)}                 & \multirow{2}{*}{10$\times$10} & TRWS   & 93.2         & 1.92         & 100\tabularnewline
                                                       &                        & D-Wave & 93.2         & 1.92         & 100\tabularnewline
      \bottomrule
    \end{tabular}
    \par\end{centering}
  \label{tbl:set3 results}
\end{table*}

\subsubsection{InSAR Images Samples} 
\begin{figure*}
\begin{centering}
\begin{tabular}{cc}
\includegraphics[width=8cm]{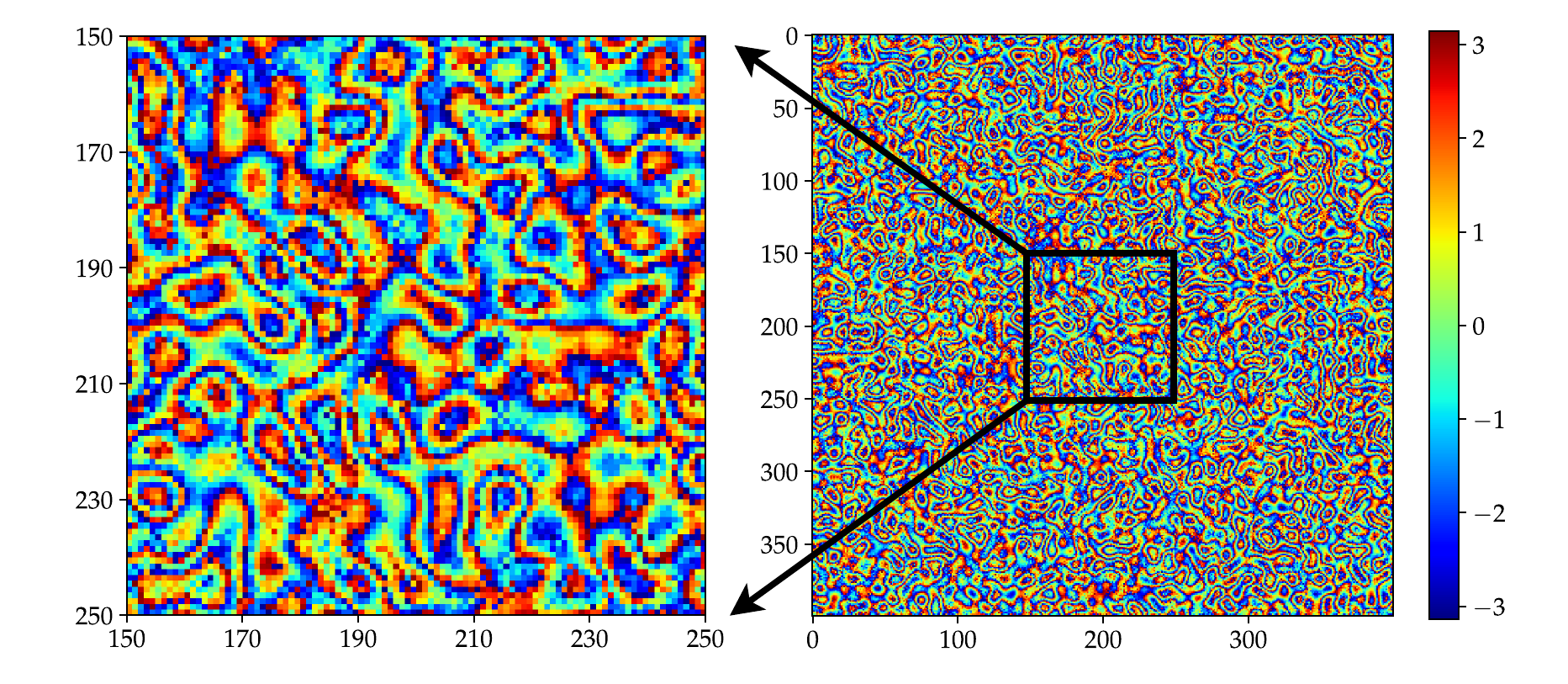} & \includegraphics[width=8cm]{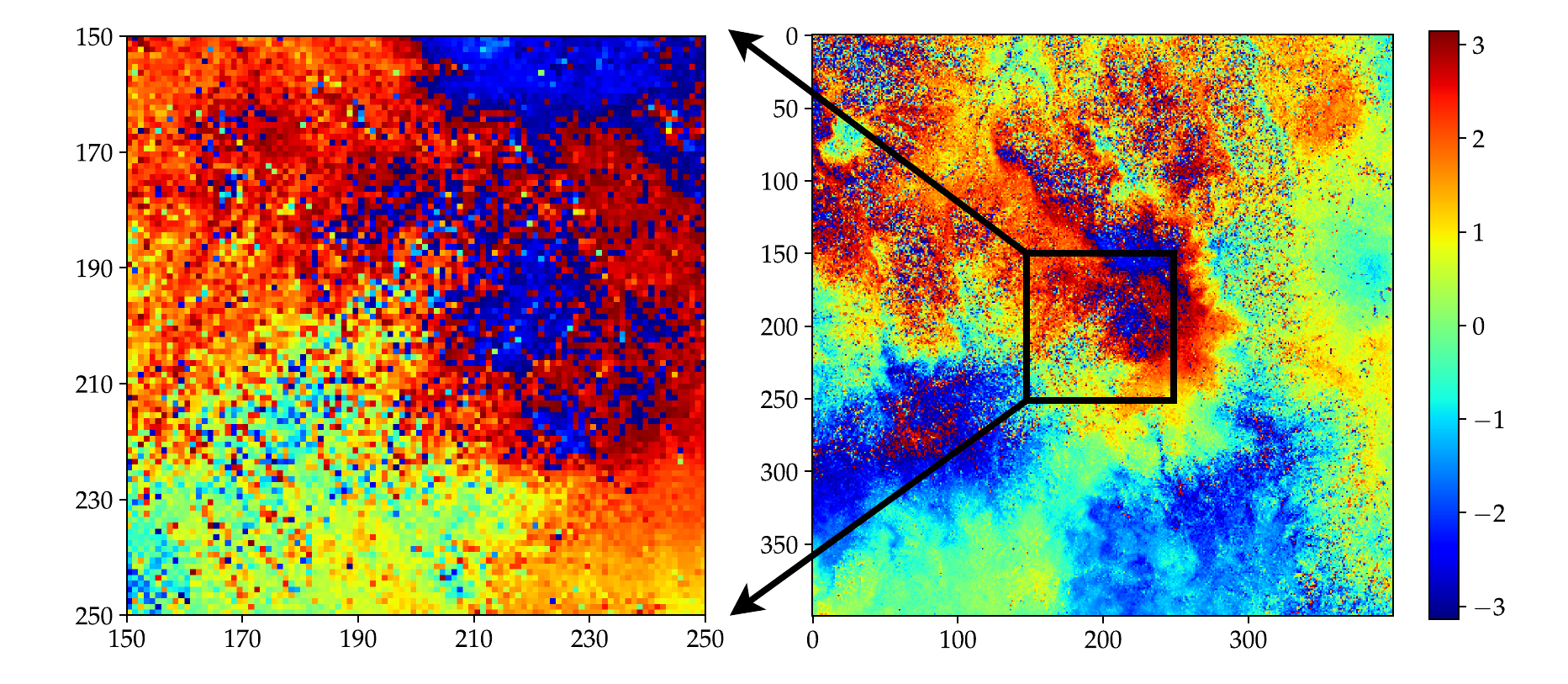}\tabularnewline
Wrapped simulated image & Wrapped real image\tabularnewline
\includegraphics[width=8cm]{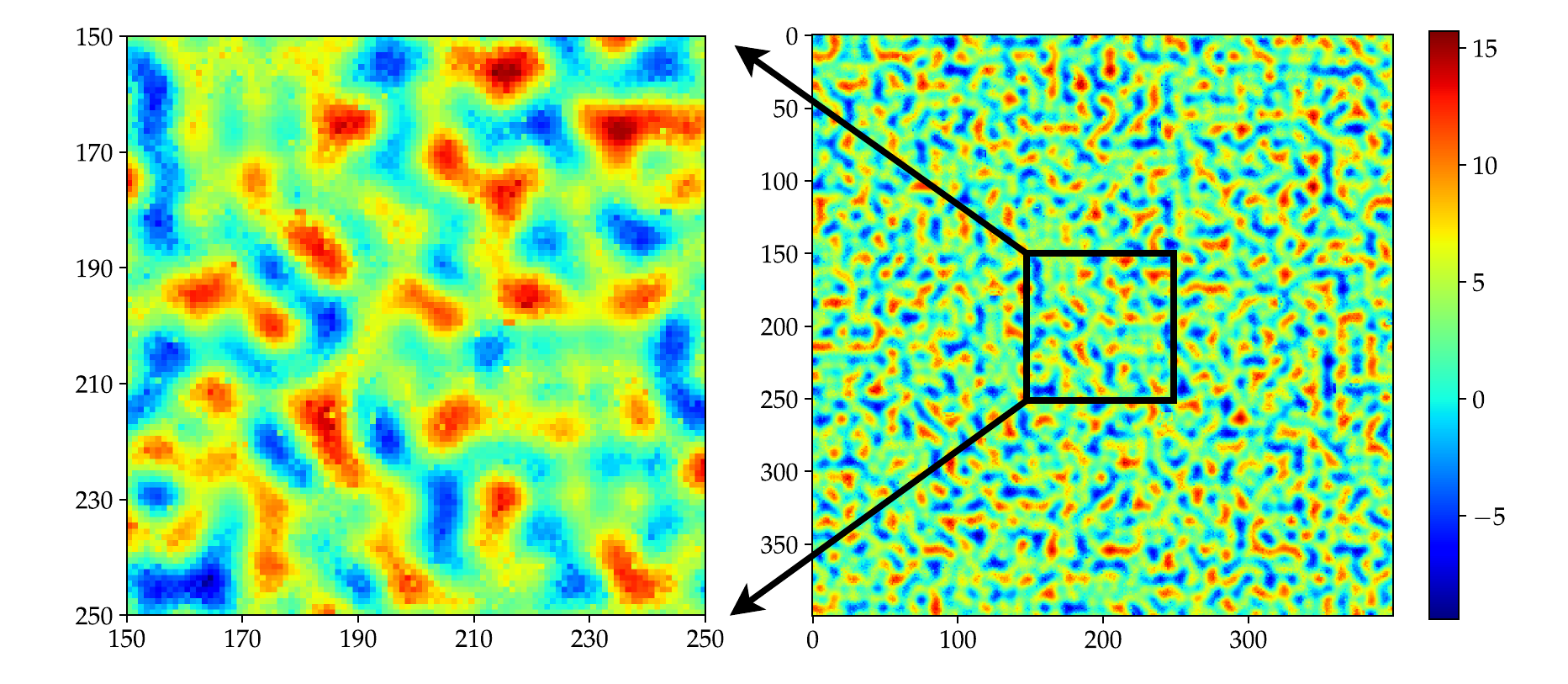} & \includegraphics[width=8cm]{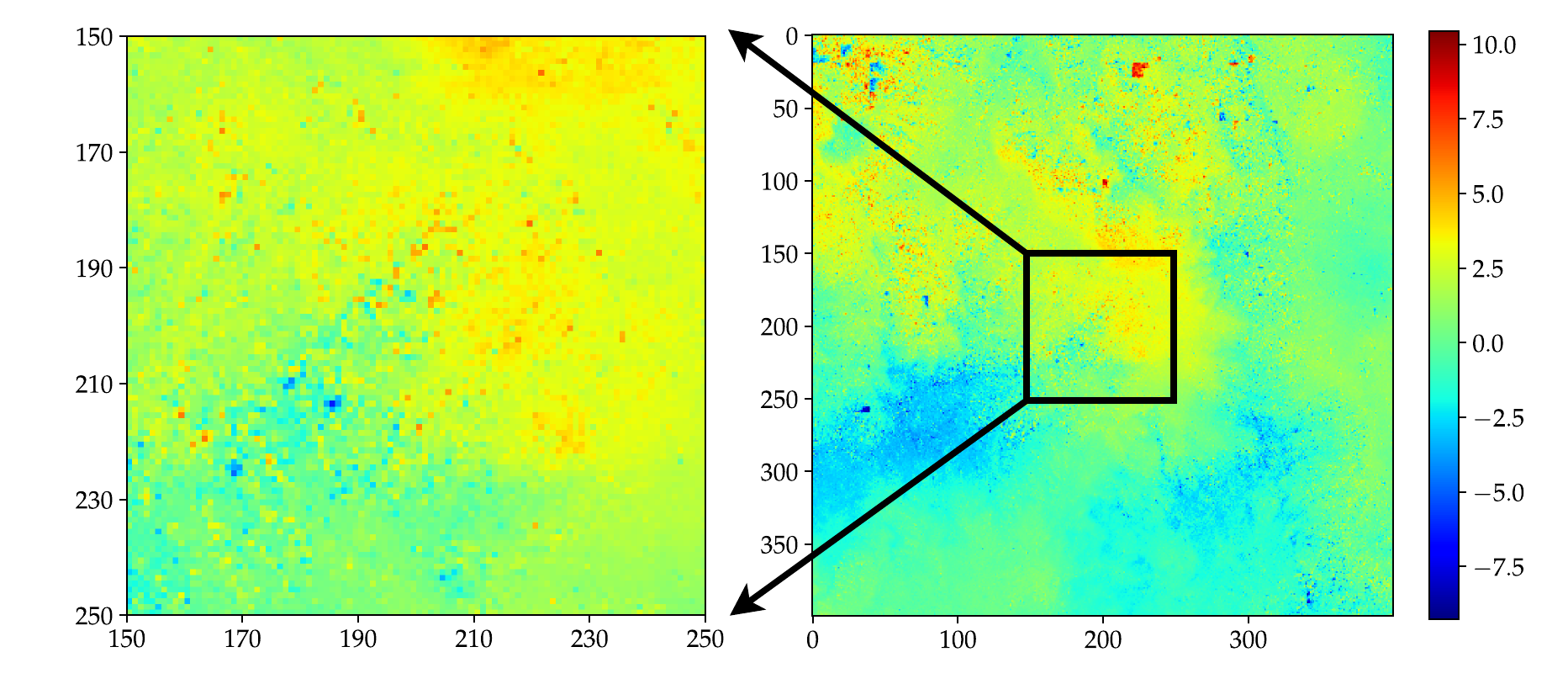}\tabularnewline
Unwrapped simulated image & Unwrapped real image\tabularnewline
\includegraphics[width=8cm]{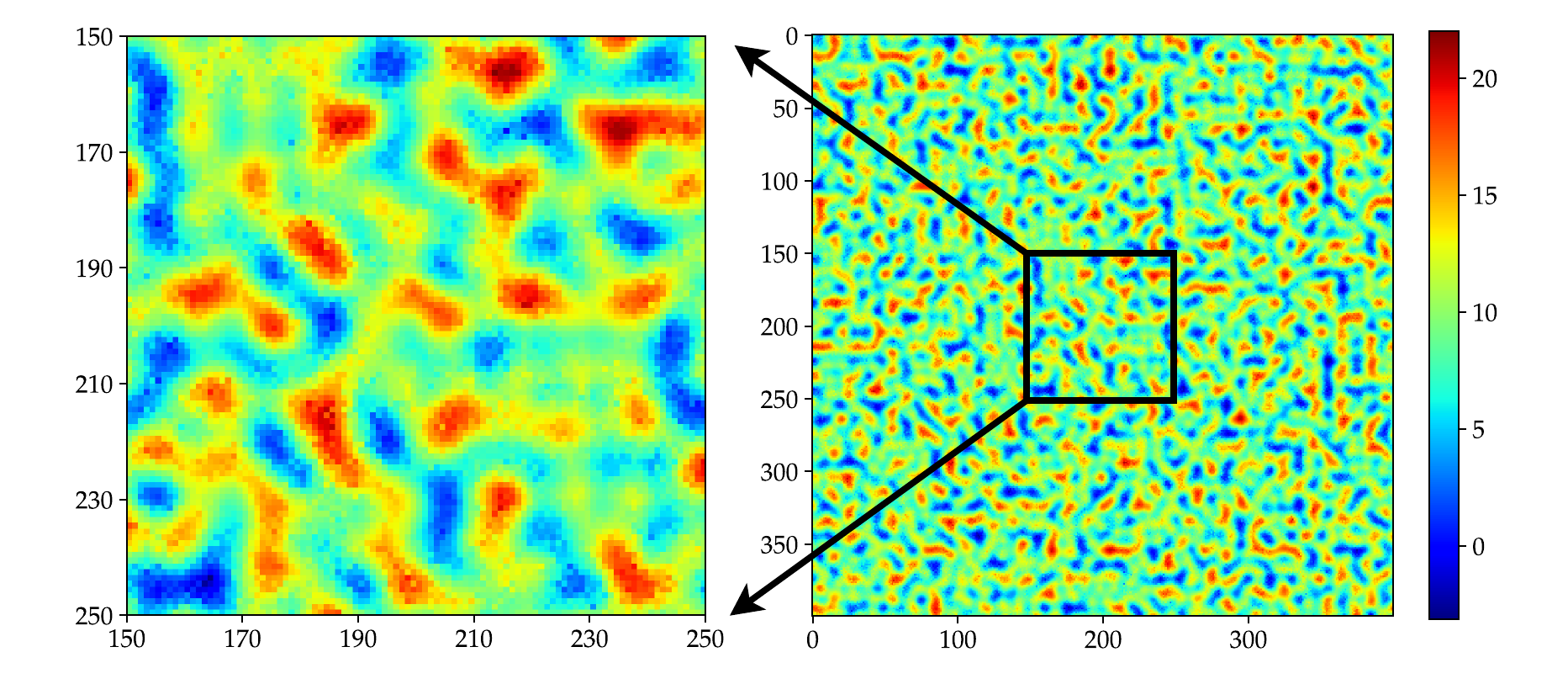} & \includegraphics[width=8cm]{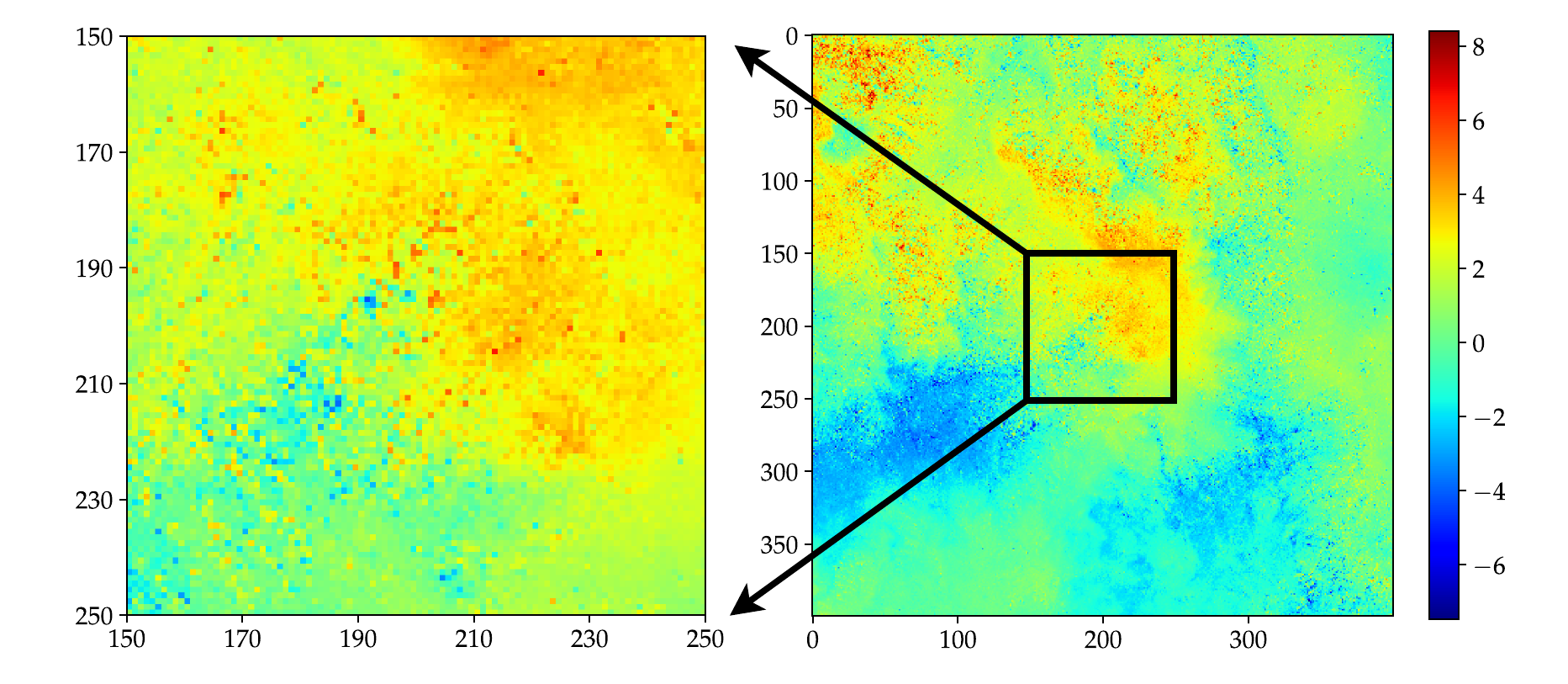}\tabularnewline
Simulated image ground truth & Real image ground truth\tabularnewline
\end{tabular}
\par
\end{centering}
\caption{Samples from the images unwrapped using the super-pixel approach. The left-hand column shows the unwrapping of a simulated high-noise image while the right-hand column shows the unwrapping of a real image. Top: the wrapped images; middle: the unwrapped images; bottom: the ground truth images to compare against.}
\par
\label{fig:insar_samples}
\end{figure*}

Fig.~\ref{fig:insar_samples} shows samples of the unwrapped images using the super-pixel method.

\subsubsection{Discussion}

The results of Experiment 1 and \mbox{Experiment 2} show that the super-pixel decomposition gives results that are very close to the ground truth, where the highest match is 100\% and the lowest is 92.78\%.
The TRWS algorithm gives the best results, followed by simulated annealing, and then parallel tempering.

Experiment 3 shows very preliminary results in using the D-Wave quantum annealing system for the phase-unwrapping problem. The results show that the phase-unwrapping problem is amenable to be solved using a real quantum annealer.
However, the quality of the solution depends greatly on the embedding.

The approach works with QUBO solvers by overcoming the problem size limitation, enabling the capability to use a quantum annealing system to unwrap large InSAR images.
In addition, it can be parallelized, which can speed up the unwrapping process.

\section{Conclusion}\label{section:Conclusion}
We formulated the phase-unwrapping problem of InSAR imaging as a QUBO problem, which we solved using a variety of QUBO solvers.
To deal with large images, we devised a divide-and-conquer method, in which a large image is decomposed into smaller images for which the phase-unwrapping problem is solved individually. We then derived another QUBO formulation based on these individual solutions. The solving of this QUBO problem yields additive integer factors that, when applied to the individual solutions, gives a solution that is very close to the global optimum.

We have tested our approach on a variety of software-implemented QUBO solvers and the D-Wave 2000Q annealer, and for a variety of both synthetic and real images. The solutions derived by our method are either identical to the ground truth or have less than 5\% of pixels differing from the ground truth.
The accuracy of the results is critically dependent on the specific annealer employed, as annealers yield “good” suboptimal solutions.

The main complexity of our experiments with the D-Wave 2000Q annealer was the mapping of the binary variables of our problem to the machine’s physical qubits. The machine has only limited connectivity between qubits. If the geometry of the problem does not match that of the machine itself, chains of qubits must be formed to ensure that the appropriate topology is achieved.
In our experiments, the results we obtained differed from the ground truth by a maximum of 15\% of the pixels. 
This lower-quality result was observed when large images were processed, which necessitated the use of all the  qubit cells in the machine. As such, the quality of the solution was affected by the faulty qubits and the asymmetries introduced in bypassing them. When smaller images were involved, which enabled the non-involvement of the faulty cells, the quality of the solutions improved drastically, reaching from 93.2\% to 100\% of the optimum.

In conclusion, we have successfully demonstrated that the SAR phase-unwrapping problem can be expressed and solved as a QUBO problem. By partitioning the problem, we have been able to obtain high-quality solutions for large images. The QUBO solvers of the Microsoft QIO toolkit achieve state-of-the-art solution quality. 

Looking ahead, our results suggest that quantum annealers may have promise as fast solvers for the phase-unwrapping problem. 
The high quality of solutions obtained using a quantum annealer leads us to believe that similarly high-quality results could be obtained for higher levels of ambiguity and larger neighbourhoods, with upcoming improvements to the architecture and qubit connectivity of future quantum annealers.

It is our intention to quantify the quality of the solutions obtained and its dependence on the size of the partitions, along with the complexity and noise content of the images we process.
We intend to study this by using both simulations and real hardware, especially the forthcoming generation of quantum annealers.

\bibliographystyle{plain}
\bibliography{citations}

\begin{thebibliography}{10}

\bibitem{DWaveQPU21:online}
{D-Wave QPU Architecture: Chimera}.
\newblock \url{https://docs.dwavesys.com/docs/latest/c_gs_4.html}.
\newblock D-Wave System Documentation.

\bibitem{chen2002phase}
Curtis~W Chen and Howard~A Zebker.
\newblock Phase unwrapping for large sar interferograms: Statistical
  segmentation and generalized network models.
\newblock {\em IEEE Transactions on Geoscience and Remote Sensing},
  40(8):1709--1719, 2002.

\bibitem{costantini1998novel}
Mario Costantini.
\newblock A novel phase unwrapping method based on network programming.
\newblock {\em IEEE Transactions on geoscience and remote sensing},
  36(3):813--821, 1998.

\bibitem{curlander1991synthetic}
John~C Curlander and Robert~N McDonough.
\newblock Synthetic aperture radar- systems and signal processing(book).
\newblock {\em New York: John Wiley \& Sons, Inc, 1991.}, 1991.

\bibitem{finnila1994quantum}
Aleta~Berk Finnila, MA~Gomez, C~Sebenik, Catherine Stenson, and Jimmie~D Doll.
\newblock Quantum annealing: a new method for minimizing multidimensional
  functions.
\newblock {\em Chemical physics letters}, 219(5-6):343--348, 1994.

\bibitem{geman1984stochastic}
Stuart Geman and Donald Geman.
\newblock Stochastic relaxation, gibbs distributions, and the bayesian
  restoration of images.
\newblock {\em IEEE Transactions on pattern analysis and machine intelligence},
  (6):721--741, 1984.

\bibitem{glover2018tutorial}
Fred Glover, Gary Kochenberger, and Yu~Du.
\newblock A tutorial on formulating and using qubo models.
\newblock {\em arXiv preprint arXiv:1811.11538}, 2018.

\bibitem{heim2015quantum}
Bettina Heim, Troels~F R{\o}nnow, Sergei~V Isakov, and Matthias Troyer.
\newblock Quantum versus classical annealing of ising spin glasses.
\newblock {\em Science}, 348(6231):215--217, 2015.

\bibitem{johnson2011quantum}
Mark~W Johnson, Mohammad~HS Amin, Suzanne Gildert, Trevor Lanting, Firas Hamze,
  Neil Dickson, Richard Harris, Andrew~J Berkley, Jan Johansson, Paul Bunyk,
  et~al.
\newblock Quantum annealing with manufactured spins.
\newblock {\em Nature}, 473(7346):194--198, 2011.

\bibitem{jooya2017accelerating}
Ali Jooya, Babak Keshavarz, Nikitas Dimopoulos, and Jaspreet~S Oberoi.
\newblock Accelerating neural network ensemble learning using optimization and
  quantum annealing techniques.
\newblock In {\em Proceedings of the Second International Workshop on Post
  Moores Era Supercomputing}, pages 1--7, 2017.

\bibitem{kolmogorov2006convergent}
Vladimir Kolmogorov.
\newblock Convergent tree-reweighted message passing for energy minimization.
\newblock {\em IEEE transactions on pattern analysis and machine intelligence},
  28(10):1568--1583, 2006.

\bibitem{nielsen2002quantum}
Michael~A Nielsen and Isaac Chuang.
\newblock Quantum computation and quantum information, 2002.

\bibitem{swendsen1986replica}
Robert~H Swendsen and Jian-Sheng Wang.
\newblock Replica monte carlo simulation of spin-glasses.
\newblock {\em Physical review letters}, 57(21):2607, 1986.

\bibitem{zaribafiyan2017systematic}
Arman Zaribafiyan, Dominic~JJ Marchand, and Seyed Saeed~Changiz Rezaei.
\newblock Systematic and deterministic graph minor embedding for cartesian
  products of graphs.
\newblock {\em Quantum Information Processing}, 16(5):136, 2017.

\bibitem{zhu2020borealis}
Zheng Zhu, Chao Fang, and Helmut~G Katzgraber.
\newblock borealis—a generalized global update algorithm for boolean
  optimization problems.
\newblock {\em Optimization Letters}, pages 1--20, 2020.

\end{thebibliography}

\clearpage
\appendices

\section{Cost Derivation}
\label{FirstAppendix}
Denoting by $\varphi_{i}$ the phase of pixel $i$, and by $\phi_{i}$
the wrapped phase of the same pixel, we can relate the phase and wrapped
phases of pixels $i$ and $j$ as follows.

\begin{equation}
\varphi_{i}=\phi_{i}+2\pi k_{i}
\label{eq:a1}
\end{equation}

and

\begin{equation}
\varphi_{j}=\phi_{j}+2\pi k_{j}
\label{eq:a2}
\end{equation}

Further, due to the Nyquist criterion, and if pixels $i$ and $j$ are neighbouring, then

\begin{equation}
\varphi_{i}-\varphi_{j}<2\pi\,.
\label{eq:Nyquist}
\end{equation}

The wrap function is defined as

\begin{equation}
\text{wrap}\left(\theta\right)=\text{arg}\left(e^{i\theta}\right)=\theta-\lfloor\frac{\theta}{2\pi}\rfloor\,.
\end{equation}

Then, we can reason as follows: from \eqref{eq:a1} and \eqref{eq:a2}, we have

\begin{equation}
\varphi_{i}-\varphi=\phi_{i}-\phi_{j}+2\pi\left(k_{i}-k_{j}\right)\,,
\label{eq:a3}
\end{equation}

\noindent
or, applying the $\text{wrap}(.)$ function on both sides, we obtain

\[
\text{wrap}\left(\varphi_{i}\text{-}\varphi_{j}\right)=\text{wrap}\left(\text{\ensuremath{\phi}}_{i}\text{-\ensuremath{\phi}}_{j}+2\pi\left(k_{i}-k_{j}\right)\right)\Rightarrow
\]

\begin{dmath}
\varphi_{i}-\varphi_{j}-2\pi\lfloor\frac{\varphi_{i}-\varphi_{j}}{2\pi}\rfloor=\phi_{i}-\phi_{j}+2\pi\left(k_{i}-k_{j}\right)-2\pi\lfloor\frac{\phi_{i}-\phi_{j}+2\pi\left(k_{i}-k_{j}\right)}{2\pi}\rfloor\,.
\label{eq:a4}
\end{dmath}

Because of the Nyquist assumption (c.f. \eqref{eq:Nyquist}),

\[
\lfloor\frac{\varphi_{i}-\varphi_{j}}{2\pi}\rfloor=0
\]

\noindent
and therefore equation \eqref{eq:a4} can be written as

\vspace{1.2em}
$
\varphi_{i}-\varphi_{j}=\phi_{i}-\phi_{j}+2\pi\left(k_{i}-k_{j}\right)-2\pi\lfloor\frac{\phi_{i}-\phi_{j}+2\pi\left(k_{i}-k_{j}\right)}{2\pi}\rfloor\Rightarrow
$

\[
\varphi_{i}-\varphi_{j}=\phi_{i}-\phi_{j}+2\pi\left(k_{i}-k_{j}\right)-2\pi\lfloor\frac{\phi_{i}-\phi_{j}}{2\pi}+\left(k_{i}-k_{j}\right)\rfloor\Rightarrow
\]

\[
\varphi_{i}-\varphi_{j}=\phi_{i}-\phi_{j}+2\pi\left(k_{i}-k_{j}\right)-2\pi\lfloor\frac{\phi_{i}-\phi_{j}}{2\pi}\rfloor-2\pi\left(k_{i}-k_{j}\right),
\]

\vspace{1.2em}

\noindent
as $\left(k_{i}-k_{j}\right)$ is an integer. Therefore,

\begin{equation}
\varphi_{i}-\varphi_{j}=\phi_{i}-\phi_{j}-2\pi\lfloor\frac{\phi_{i}-\phi_{j}}{2\pi}\rfloor=\text{wrap}\left(\phi_{i}-\phi_{j}\right).
\label{eq:a5}
\end{equation}

\vspace{1.2em}

Using equation \eqref{eq:a5} and \eqref{eq:a3}, we obtain 

\[
\varphi_{i}-\varphi_{j}=\phi_{i}-\phi_{j}+2\pi\lfloor k_{i}-k_{j}\rfloor=\text{wrap}\left(\phi_{i}-\phi_{j}\right)\Rightarrow
\]

\begin{equation}
k_{i}-k_{j}=\frac{\text{wrap}\left(\phi_{i}-\phi_{j}\right)-\left(\phi_{i}-\phi_{j}\right)}{2\pi}.
\label{eq:a6}
\end{equation}

\vspace{0.8em}

Denoting

\begin{equation}
a_{ij}\stackrel{\text{def}}{=}\frac{\text{wrap}\left(\phi_{i}-\phi_{j}\right)-\left(\phi_{i}-\phi_{j}\right)}{2\pi},
\end{equation}

\noindent
equation \eqref{eq:a6} is written as 

\begin{equation}
k_{i}-k_{j}=a_{ij}\Rightarrow k_{i}-k_{j}-a_{ij}=0.
\end{equation}

\vspace{0.8em}

This equation is the basis of the cost function the optimization of
which will produce appropriate values for the labels $k_{i}$.

The unwrapping problem can then be expressed as an optimization problem
of the cost function 
\begin{equation}
E=\sum_{(s,t)\text{\ensuremath{\in}}A}W_{st}|k_{t}-k_{s}-a_{st}|\,,
\end{equation}
that is, 
\begin{equation}
argmin_{k}=\sum_{(s,t)\text{\ensuremath{\in}}A}W_{st}|k_{t}-k_{s}-a_{st}|\,,
\end{equation}
where $k_{i}$ are the labels that will determine the original phase
as per Equation (1), $A$ is the set of pixels in the SAR image, and $W_{st}$
are weights defining the neighbourhood structure.

\end{document}